# Speckle Noise Analysis for Synthetic Aperture Radar (SAR) Space Data

Sanjjushri Varshini R (sanjjushrivarshini@gmail.com), Rohith Mahadevan (mailrohithmahadevan@gmail.com), Bagiya Lakshmi S(bagiyalakshmi59@gmail.com ), Mathivanan Periasamy (mathivananperiasamy@gmail.com), Raja CSP Raman (raja.csp@gmail.com), Lokesh M (lokeshshivalover@gmail.com)

## Abstract

This research tackles the challenge of speckle noise in Synthetic Aperture Radar (SAR) space data, a prevalent issue that hampers the clarity and utility of SAR images. The study presents a comparative analysis of six distinct speckle noise reduction techniques: Lee Filtering, Frost Filtering, Kuan Filtering, Gaussian Filtering, Median Filtering, and Bilateral Filtering. These methods, selected for their unique approaches to noise reduction and image preservation, were applied to SAR datasets sourced from the Alaska Satellite Facility (ASF). The performance of each technique was evaluated using a comprehensive set of metrics, including Peak Signal-to-Noise Ratio (PSNR), Mean Squared Error (MSE), Structural Similarity Index (SSIM), Equivalent Number of Looks (ENL), and Speckle Suppression Index (SSI). The study concludes that both the Lee and Kuan Filters are effective, with the choice of filter depending on the specific application requirements for image quality and noise suppression. This work provides valuable insights into optimizing SAR image processing, with significant implications for remote sensing, environmental monitoring, and geological surveying.

## 1 Introduction

This research addresses the persistent issue of speckle noise in Synthetic Aperture Radar (SAR) space data. Speckle noise, an inherent granular noise that degrades the quality of SAR images, poses considerable challenges in accurately interpreting and analyzing these images. Byeffectively reducing speckle noise, the clarity and usability of SAR data can be greatly enhanced, fostering improved applications in various fields such as remote sensing, environmental monitoring, and geological surveying.

This research delivers a remarkable and innovative strategy by employing six distinct techniques for speckle noise reduction. Unlike established methods that rely on a single filtering technique, this study explores and compares multiple approaches to determine the most effective strategy. Including various filters allows for a comprehensive analysis and offers a more robust solution to the problem of speckle noise, exhibiting a novel methodology in the domain of SAR image processing.

The techniques investigated in this study include Lee Filtering, Frost Filtering, Kuan Filtering, Gaussian Filtering, Median Filtering, and Bilateral Filtering. Each of these methods has its own set of characteristics and operational principles, providing diverse options for speckle noise reduction. Lee Filtering and Frost Filtering are known for their adaptive nature, while Kuan Filtering is valued for its statistical approach. Gaussian Filtering, with its smoothing properties, Median Filtering, known for edge preservation, and Bilateral Filtering, which combines both spatial and intensity domain filtering, add further dimensions to the analysis.

The methodology adopted in this study involves the application of these six filters to SAR space data, followed by a comparative analysis of their performance. By systematically assessing the effectiveness of each filter, the study aims to identify the most appropriate technique or combination of methods for optimal speckle noise reduction. This comprehensive approach advances the understanding of speckle noise mitigation and provides practical insights for enhancing the quality of SAR Space imagery.

## 2 Literature Review

The summary of the most cited articles on SAR data analysis found in the Mendeley database here.

Cloude and Pottier [1] present a method for parameterizing polarimetric scattering problems using eigenvalue analysis of the coherency matrix. This method applies a three-level Bernoulli statistical model to estimate average target scattering matrix parameters, emphasizing scattering entropy as a key factor in assessing polarimetric SAR data. The method is validated using POLSAR data from NASA/JPL AIRSAR and classical random media scattering problems.

Freeman and Durden [2] developed a model combining three scattering mechanisms—canopy scatter, even/double-bounce scatter, and Bragg scatter—to describe polarimetric SAR backscatter from natural scatterers. This model effectively distinguishes between various forest conditions using data from NASA/JPL's AIRSAR system and serves as a predictive tool for estimating forest inundation and disturbance effects.

Ferretti and colleagues [3] present a procedure to identify and use stable natural reflectors or permanent scatterers from long series of interferometric SAR images. Their method, applied to ESA ERS data, achieves high-accuracy DEM and terrain motion detection by estimating and removing atmospheric phase contributions, demonstrated through motion measurements and DEM refinement in Ancona, Italy.

Berardino and colleagues [4] introduce a differential SAR interferometry algorithm to monitor surface deformations over time. Using singular value decomposition, the technique links independent SAR data sets to increase temporal observation, filtering atmospheric phase artifacts. The approach, tested with European Remote Sensing satellite data from 1992 to 2000, tracks surface deformation dynamics in Campi Flegrei caldera and Naples, Italy.

Combot and colleagues [5] use spaceborne SAR to improve descriptions of air–sea exchanges under tropical cyclones. Their database, constructed from RadarSat-2 and Sentinel-1, includes high-resolution wind fields for 161 cyclones. The methodology, validated against best track and SFMR measurements, effectively captures TC vortex structure, despite challenges in heavy precipitation.

Abdel-Hamid and colleagues [6] assess drought stress on grasslands in Eastern Cape Province using Sentinel-1 SAR data. Their analysis shows a significant correlation between SAR backscattering coefficients and NDVI values. The study finds communal grasslands more affected by drought than commercial ones, highlighting the role of management in improving resilience and productivity.

Sekertekin and colleagues [7] explore the potential of ALOS-2 and Sentinel-1 SAR data for soil moisture estimation. Their analysis shows Sentinel-1 outperforms ALOS-2 for bare soil surfaces, while both data sets provide satisfactory results in vegetated surfaces using the Water Cloud Model. The study highlights the higher accuracy of Sentinel-1 for soil moisture estimation.

Mullissa and colleagues [8] propose a framework for preparing Sentinel-1 SAR backscatter data in Google Earth Engine, incorporating noise correction, speckle filtering, and radiometric terrain normalization. This framework generates Analysis-Ready-Data suitable for various land and water applications.

Zhang and colleagues [9] review the SAR Ship Detection Dataset (SSDD), emphasizing its popularity and impact on SAR remote sensing. They address limitations in initial versions by introducing bounding box, rotatable bounding box, and polygon segmentation labels, along with strict usage standards to improve accuracy and academic exchanges.

Gagliardi and colleagues [10] demonstrate the use of Sentinel-1A SAR data for monitoring runway displacements at Leonardo Da Vinci International Airport. Their geostatistical analysis compares SAR data with high-resolution COSMO-SkyMed and ground-based data, proving Sentinel-1A's effectiveness for long-term monitoring and maintenance strategies.

Morishita and Kobayashi [11] propose a method to derive 3D deformation by integrating deformation data from different sources. Their approach, validated with ALOS-2 data, successfully retrieves 3D coseismic deformation with high accuracy. The method is applicable to other SAR datasets and beneficial for disaster recovery.

Oveis and colleagues [12] review the application of convolutional neural networks (CNNs) in SAR data analysis, covering various subareas such as target recognition, land use classification, and change detection. The review highlights practical techniques like data augmentation and transfer learning, and discusses future research directions and challenges.

Tiampo and colleagues [13] compare methods for flood inundation mapping using SAR data and machine learning. They find amplitude thresholding the most effective technique, although machine learning also successfully reproduces inundation shapes. The study demonstrates high-resolution mapping's potential for emergency hazard response.

Ferguson and Gunn [14] review radar polarimetric decomposition for freshwater ice systems, discussing lake ice, river ice, and glacial systems. They recommend further development of ice models and methods to improve environmental observables extraction, highlighting areas for future research.

The reviewed literature demonstrates the diverse applications of SAR data and the challenges posed by speckle noise. Filter-based methods provide a practical and effective approach to address this issue. By carefully selecting and designing filters, significant improvements in image quality can be achieved, leading to more accurate and reliable results across a wide range of SAR applications.

# 3 Proposed Solution

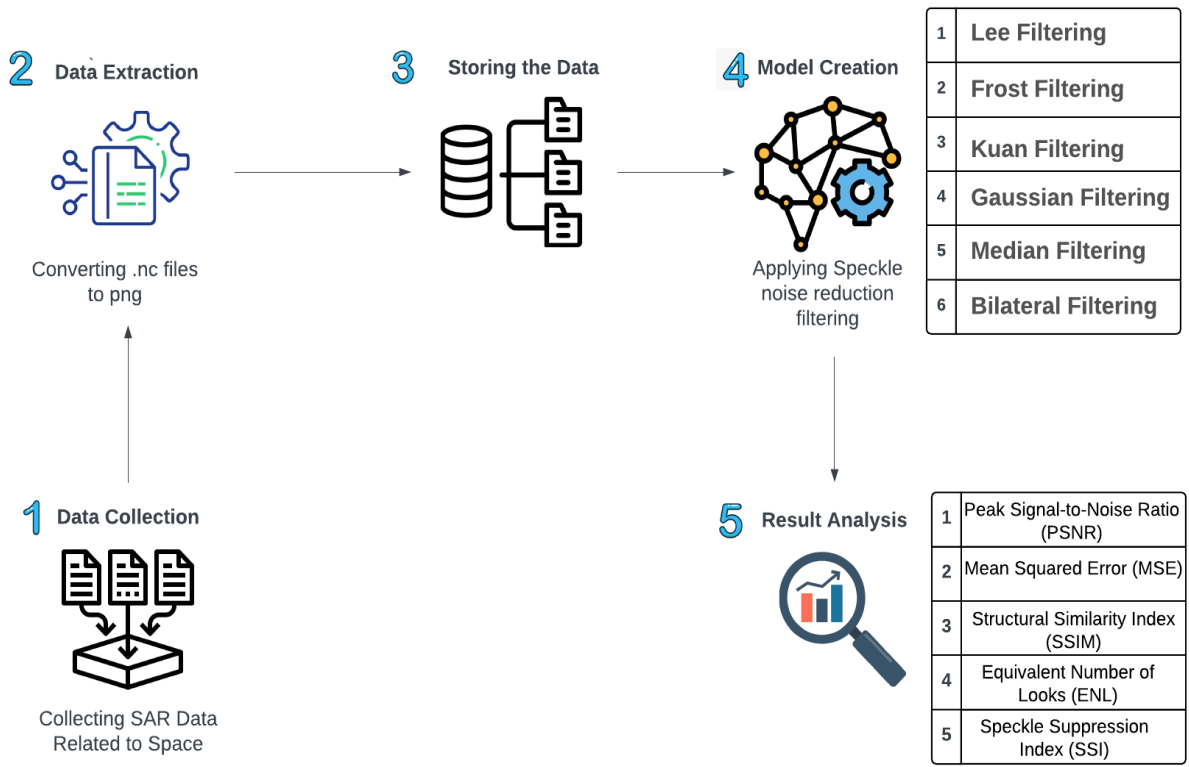

## 3.1 Data Collection

The initial step involves collecting Synthetic Aperture Radar (SAR) space data from the Alaska Satellite Facility (ASF). ASF provides a wealth of SAR data from various satellites, ensuring access to high-quality and diverse datasets. This stage is critical as the subsequent analysis's accuracy and reliability depend on the collected data's quality. The selection of datasets will be based on criteria such as resolution, coverage area, and the presence of speckle noise, which is inherent in SAR imagery.

## 3.2 Data Extraction

Once the SAR data is collected, the next step is to convert the raw data files, typically in the NetCDF (.nc) format, into a more accessible image format such as PNG. This conversion is necessary for applying image processing techniques. The process involves extracting the relevant image data from the .nc files, ensuring that the spatial and radiometric integrity of the data is maintained. Specialized software tools and libraries, such as GDAL (Geospatial Data Abstraction Library), can be utilized for this purpose, facilitating efficient and accurate conversion.

## 3.3 Data Storing

After conversion, the images are stored in a structured manner for easy access and processing. Proper metadata is also recorded to maintain the context of the images, which is essential for the accuracy of the subsequent analysis.

## 3.4 Applying Speckle Noise Reduction Techniques

The core of this research lies in applying various speckle noise reduction techniques to the SAR Space images. Each method has unique characteristics and offers different advantages:

**Lee Filtering:**
Utilizes local statistics to smooth the image while preserving edges adaptively. It is particularly effective in reducing speckle noise without compromising significant image features.

**Frost Filtering:**
Applies an exponential kernel that adapts to local variations in the image, effectively reducing noise while maintaining the integrity of edges and fine details.

**Kuan Filtering:**
Uses a multiplicative noise model to analyze and reduce speckle noise statistically. It balances between smoothing homogeneous areas and preserving edges.

**Gaussian Filtering:**
A linear filter that smooths the image based on a Gaussian function, effectively reducing noise at the cost of slight blurring of edges.

**Median Filtering:**
A non-linear filter that replaces each pixel value with the median value of the neighboring pixels. It is effective in reducing impulsive noise while preserving edges.

**Bilateral Filtering:**
Combines spatial and intensity domain filtering to smooth images while preserving edges and fine details. It is particularly effective in maintaining high-quality image post-processing.

## 3.5 Result Analysis

The final step involves a thorough analysis of the results obtained from applying the different filtering techniques.

**PSNR (Peak Signal-to-Noise Ratio)**: Measures the ratio between the maximum possible signal power and the power of corrupting noise, indicating the quality of the reconstructed image.

**MSE (Mean Squared Error)**: Calculates the average of the squares of the differences between the original and denoised image pixels, reflecting the overall error.

**SSIM (Structural Similarity Index)**: Assesses the similarity between the original and denoised images by comparing luminance, contrast, and structure, providing a perceptual quality measure.

**ENL (Equivalent Number of Looks)**: Evaluates the extent of speckle noise reduction by measuring the homogeneity of uniform image areas.

**SSI (Speckle Suppression Index)**: Quantifies the effectiveness of speckle noise reduction, balancing noise suppression and detail preservation in the image.

## 4 Result

In our research, we have evaluated various speckle noise reduction algorithms using multiple evaluation metrics. These metrics include Peak Signal-to-Noise Ratio (PSNR), Mean Squared Error (MSE), Structural Similarity Index (SSIM), Equivalent Number of Looks (ENL), and Speckle Suppression Index (SSI). The performance of five different filters—Lee, Kuan, Gaussian, Median, and Bilateral—was assessed using these metrics. The results are summarized in the table below:

|   | Metric | Lee Filter | Kuan Filter | Gaussian Filter | Median Filter | Bilateral Filter |
|---|--------|-----------|-------------|-----------------|---------------|------------------|
| 0 | PSNR   | 40.206693 | 37.984304   | 29.072036       | 29.669396     | 29.907636        |
| 1 | MSE    | 6.200277  | 10.343083   | 80.514994       | 70.168261     | 66.422740        |
| 2 | SSIM   | 0.969557  | 0.986636    | 0.659307        | 0.804169      | 0.840699         |
| 3 | ENL    | 0.633798  | 0.697551    | 0.872117        | 0.645966      | 0.689973         |
| 4 | SSI    | 1.103162  | 1.121805    | 4.188730        | 3.502629      | 2.846183         |

**PSNR**: The Lee Filter achieved the highest PSNR value of 40.206693, indicating the best performance in terms of signal preservation.

**MSE**: The Lee Filter also performed the best in terms of MSE, with the lowest value of 6.200277, suggesting minimal error.

**SSIM**: The Kuan Filter showed the highest SSIM value of 0.986636, indicating superior structural similarity and image quality.

**ENL**: The Gaussian Filter obtained the highest ENL value of 0.872117, suggesting it is the best at reducing speckle noise.

**SSI**: The Lee and Kuan Filters both demonstrated strong performance in terms of SSI, with values of 1.103162 and 1.121805, respectively, indicating effective speckle suppression.

## 5 Future scope

The review of literature highlights several critical areas where Synthetic Aperture Radar (SAR) data analysis has made significant advancements, particularly in polarimetric scattering, surface deformation monitoring, and environmental applications. However, there remains substantial room for future research and development. With the growing application of convolutional neural networks (CNNs) in SAR data analysis, future research could focus on developing more sophisticated machine learning models, such as deep learning architectures, to enhance the accuracy and efficiency of tasks like target recognition, land use classification, and change detection. Exploring the potential of hybrid models combining machine learning with traditional SAR processing techniques could also yield promising results. The integration of SAR data with other remote sensing data, such as optical imagery or LiDAR, presents an exciting opportunity for enhanced environmental monitoring and disaster management. Future studies could focus on developing frameworks that effectively merge data from different sources to provide more comprehensive and accurate assessments of natural phenomena, such as flooding, landslides, and forest conditions.

## 6 Conclusion

Overall, the Lee Filter exhibits the best performance across several metrics, making it a strong candidate for speckle noise reduction. However, the Kuan Filter emerges as a viable alternative, offering a good balance with high PSNR, low MSE, high SSIM, and reasonable ENL. While the Lee Filter provides slightly better image detail preservation, the Kuan Filter's performance in multiple metrics suggests it could be the preferred choice for effective speckle noise reduction.

# 7 References

[1]     S. R. Cloude and E. Pottier, "An entropy based classification scheme for land applications of polarimetric SAR," IEEE Transactions on Geoscience and Remote Sensing, vol. 35, no. 1, 1997, doi: 10.1109/36.551935.

[2]     A. Freeman and S. L. Durden, "A three-component scattering model for polarimetric SAR data," IEEE Transactions on Geoscience and Remote Sensing, vol. 36, no. 3, 1998, doi: 10.1109/36.673687.

[3]     A. Ferretti, C. Prati, and F. Rocca, "Permanent scatterers in SAR interferometry," IEEE Transactions on Geoscience and Remote Sensing, vol. 39, no. 1, 2001, doi: 10.1109/36.898661.

[4]     P. Berardino, G. Fornaro, R. Lanari, and E. Sansosti, "A new algorithm for surface deformation monitoring based on small baseline differential SAR interferograms," IEEE Transactions on Geoscience and Remote Sensing, vol. 40, no. 11, 2002, doi: 10.1109/TGRS.2002.803792.

[5]     C. Combot et al., "Extensive high-resolution synthetic aperture radar (SAR) data analysis of tropical cyclones: Comparisons with SFMR flights and best track," Mon Weather Rev, vol. 148, no. 11, 2020, doi: 10.1175/MWR-D-20-0005.1.

[6]     A. Abdel-Hamid, O. Dubovyk, V. Graw, and K. Greve, "Assessing the impact of drought stress on grasslands using multi-temporal SAR data of Sentinel-1: a case study in Eastern Cape, South Africa," Eur J Remote Sens, vol. 53, no. sup2, 2020, doi: 10.1080/22797254.2020.1762514.

[7]     A. Sekertekin, A. M. Marangoz, and S. Abdikan, "ALOS-2 and Sentinel-1 SAR data sensitivity analysis to surface soil moisture over bare and vegetated agricultural fields," Comput Electron Agric, vol. 171, 2020, doi: 10.1016/j.compag.2020.105303.

[8]     A. Mullissa et al., "Sentinel-1 sar backscatter analysis ready data preparation in google earth engine," Remote Sens (Basel), vol. 13, no. 10, 2021, doi: 10.3390/rs13101954.

[9]     T. Zhang et al., "SAR ship detection dataset (SSDD): Official release and comprehensive data analysis," Remote Sens (Basel), vol. 13, no. 18, 2021, doi: 10.3390/rs13183690.

[10]    V. Gagliardi et al., "Testing sentinel-1 sar interferometry data for airport runway monitoring: A geostatistical analysis," Sensors, vol. 21, no. 17, 2021, doi: 10.3390/s21175769.

[11]	Y. Morishita and T. Kobayashi, "Three-dimensional deformation and its uncertainty derived by integrating multiple SAR data analysis methods," Earth, Planets and Space, vol. 74, no. 1, 2022, doi: 10.1186/s40623-022-01571-z.

[12]	A. H. Oveis, E. Giusti, S. Ghio, and M. Martorella, "A Survey on the Applications of Convolutional Neural Networks for Synthetic Aperture Radar: Recent Advances," IEEE Aerospace and Electronic Systems Magazine, vol. 37, no. 5, 2022, doi: 10.1109/MAES.2021.3117369.

[13]	K. F. Tiampo, L. Huang, C. Simmons, C. Woods, and M. T. Glasscoe, "Detection of Flood Extent Using Sentinel-1A/B Synthetic Aperture Radar: An Application for Hurricane Harvey, Houston, TX," Remote Sens (Basel), vol. 14, no. 9, 2022, doi: 10.3390/rs14092261.

[14]	J. E. Ferguson and G. E. Gunn, "Polarimetric decomposition of microwave-band freshwater ice SAR data: Review, analysis, and future directions," Remote Sens Environ, vol. 280, 2022, doi: 10.1016/j.rse.2022.113176.